\PassOptionsToPackage{table}{xcolor}
\documentclass[sigconf=true, nonacm=true]{acmart}
\usepackage{booktabs} 

\copyrightyear{2021}
\acmYear{2021}
\setcopyright{none}
\acmConference[GECCO '21]{Genetic and Evolutionary Computation Conference}{July 10--14, 2021}{Lille, France}

\usepackage{amsmath}
\usepackage{xspace}
\usepackage{xcolor}
\usepackage{dsfont}
\usepackage{rotating}
\usepackage{makecell}

\usepackage{amsthm}
\usepackage{amsmath}
\renewcommand{\epsilon}{\varepsilon}

\newcommand{\AUC}{\ensuremath{\operatorname{AUC}}\xspace}
\newcommand{\AOC}{\ensuremath{\operatorname{AOC}}\xspace}

\begin{document}
\title[Assessing the Benefits of New Algorithmic Ideas via Tuning]{Tuning as a Means of Assessing the Benefits of New Ideas in Interplay with Existing Algorithmic Modules}

\author{Jacob de Nobel}
\affiliation{
  \institution{Leiden Institute of Advanced Computer Science}
  \city{Leiden}
  \country{The Netherlands}
}

\author{Diederick Vermetten}\thanks{Equal contribution between first and second authors}
\affiliation{
  \institution{Leiden Institute of Advanced Computer Science}
  \city{Leiden}
  \country{The Netherlands}
}

\author{Hao Wang}
\affiliation{
  \institution{Leiden Institute of Advanced Computer Science}
  \city{Leiden}
  \country{The Netherlands}
}

\author{Carola Doerr}
\affiliation{
  \institution{Sorbonne Universit\'e, CNRS, LIP6}
  \city{Paris}
  \country{France}}

\author{Thomas B{\"a}ck}
\affiliation{
  \institution{Leiden Institute of Advanced Computer Science}
  \city{Leiden}
  \country{The Netherlands}
}

\begin{abstract}
Introducing new algorithmic ideas is a key part of the continuous improvement of existing optimization algorithms. However, when introducing a new component into an existing algorithm, assessing its potential benefits is a challenging task. Often, the component is added to a default implementation of the underlying algorithm and compared against a limited set of other variants. This assessment ignores any potential interplay with other algorithmic ideas that share the same base algorithm, which is critical in understanding the exact contributions being made. 

We introduce a more extensive procedure, which uses hyperparameter tuning as a means of assessing the benefits of new algorithmic components. This allows for a more robust analysis by not only focusing on the impact on performance, but also by investigating how this performance is achieved. 

We implement our suggestion in the context of the Modular CMA-ES framework, which was redesigned and extended to include some new modules and several new options for existing modules, mostly focused on the step-size adaptation method. Our analysis highlights the differences between these new modules, and identifies the situations in which they have the largest contribution. 
\end{abstract}

\maketitle

\section{Introduction}
With the continuous increase in interest for the field of optimization, many new algorithms get introduced every year. A large number of these algorithms are not completely novel, but instead add new algorithmic ideas to existing methods. 
Originally referring to one particular algorithm, CMA-ES has developed into a whole family of algorithms that are built around the core design of the original CMA-ES algorithm from~\cite{hansen2001self_adaptation_es}. 
While this growth of the algorithm set helps to keep improving the state-of-the-art performance, it also 
raises a simple question: ``How to assess the benefits of new algorithmic ideas?''

The naive way of performing such an assessment is to implement the algorithmic idea into a bare-bones version of the base algorithm, and to benchmark it against the default (and maybe some other variants). While this technique does manage to give an indication of the usefulness of the newly introduced component, the results are not always practical and hide important information, since they only consider the idea in isolation. Often, there tends to be an important interplay between algorithmic components, which is completely missed when doing the type of assessment described above. 

We aim to provide in this work a roadmap for assessing these algorithmic ideas in a way which takes component interactions into account. This is achieved by considering the different algorithmic ideas as modules in a modular framework. Several of these types of frameworks have been developed over the years~\cite{XuHHL12,tavares_evolution_2004,lourenco_evolving_2012,Baretti,van_rijn_evolving_2016}. We will work here with the Modular CMA-ES (ModCMA), which is significantly extended from the existing ModEA framework~\cite{van_rijn_evolving_2016}, by both adding new modules and new options for existing modules (see Section~\ref{redModea} for details). 

With this modular framework, we show in this work how hyperparameter tuning can be used to assess the contributions of the newly implemented components. 
We illustrate how this approach gives a detailed perspective on the benefits of new algorithmic ideas, by not only looking at pure performance metrics, but also considering the interplay with existing modules. We show, among other things, that the introduction of new step-size adaptation methods can be beneficial, but that it requires careful consideration of the interactions with other modules, such as the recombination weights. We also discuss the limitations of this approach, and how to best use it to gain the most understanding about these new algorithmic ideas.

\section{Redesigning ModEA to a Modular CMA-ES Framework}
\label{redModea}
Our work relies heavily on the Modular Evolutionary Algorithms (ModEA) framework introduced in~\cite{van_rijn_evolving_2016}. Since this framework hasn't undergone any active development in recent time, we decided to redesign the framework to our specifications. The modifications we made rendered the name of the framework no longer befitting, as only CMA-ES variants can now be created using the framework, whereas the original  framework also supported the design of other evolutionary algorithms. The new framework was dubbed the Modular CMA-ES (ModCMA) and is available as an open source Python package within the IOHprofiler~\cite{IOHprofiler} environment\footnote{\url{https://github.com/IOHprofiler/ModularCMAES}}
. It is integrated with the IOHexperimenter, giving access to a broad set of benchmark problems, including a C++ implementation of the BBOB functions~\cite{bbobfunctions} from the COCO environment~\cite{hansen2020coco}. In addition, this allows for easy data logging, which can be used directly with the interactive performance analysis and visualization from IOHanalyzer~\cite{IOHanalyzer}. 

\paragraph{Motivation}
The primary goal behind redesigning the framework was to reduce its complexity, and to only include functionality compatible to the CMA-ES and its variants. The reasoning behind this is the fact that the framework mostly revolves around the CMA-ES. Other EAs are available in the framework, but are quite underdeveloped w.r.t.~the CMA-ES. Moreover, introducing working interactions between the CMA-ES and operators from other EAs overly complicates the framework's structure. For example, ModEA contains a range of different methods for performing recombination. However, the canonical CMA-ES does not explicitly perform recombination. Instead, it updates its mean $m$ by taking a weighted average of the individuals in its current population, which it then uses to sample new individuals from a normal distribution. In other EAs, recombination occurs in a much more pronounced sense, for example by crossover. In order to make the modular algorithm of the CMA-ES function with these other forms of recombination, its original method for ``recombination'' had to be adapted. The CMA-ES however, is still only able to properly function with one of these recombination methods, the canonical one. As this pattern could be observed in other parts of the framework as well (i.e., mutation, selection), it was decided to remove these other methods all together and to focus solely on the CMA-ES.

\subsection{The Modular CMA-ES}\label{sec:modcma}

To design the Modular CMA-ES, we use the implementation from the popular CMA-ES tutorial~\cite{cmatut} as a starting point. This work provides a detailed description of the CMA-ES algorithm, including a practical guide to its implementation. From this basic design, we separate the CMA-ES in a number of functionally related blocks, in order to allow a customization of a specific part of the algorithm. This allows us to implement algorithmic variants of the CMA-ES as functional modules. From a user perspective, any of these modules could then be combined in order create a custom instantiation of the CMA-ES, by selecting an option for each available module. 

In ModEA, eleven of such modules were already implemented. These were all reimplemented in the Modular CMA-ES, with a few changes to the structure of the options. Specifically, we removed the \textit{Pairwise Selection} as a module.Instead, we incorporated this option in the \textit{Mirrored Sampling} module as the option \textit{Mirrored Sampling with Pairwise Selection}, converting this module from binary to ternary. This is done because the pairwise selection method is not suited for use without mirrored sampling~\cite{auger_mirrored_2011}. 

We implemented a new module for performing boundary correction (see Section~\ref{sec:boundcorr}), and added five alternative options for performing step size adaptation (see Section~\ref{subsec:ssa}). These two extensions to the framework will be the focus of our analysis through out this work. This set of changes give us the following list of modules for the redesigned Modular CMA-ES:

\begin{enumerate}
    \item \textbf{Active Update}: Bad candidate solutions are penalized in the covariance matrix update using negative weights~\cite{jastrebski_improving_2006}. Note that in ~\cite{cmatut}, this is given as the default version, here we consider it to be optional. 
    \item \textbf{Elitism}: $(\mu + \lambda)$ - selection instead of $(\mu, \lambda)$ - selection. 
    \item \textbf{Orthogonal Sampling}: All the newly sampled points in the population are orthonormalized using a Gram-Schmidt procedure~\cite{wang_mirrored_2014}.
    \item \textbf{Sequential Selection}: Candidate solution are immediately ranked and compared with the current best solution. If improvement is found, no additional objective function evaluations are performed~\cite{brockhoff_mirrored_2010}. 
    \item \textbf{Threshold Convergence}: A method for balancing exploration with exploitation, scaling the mutation vectors to a required length threshold, which decays over time~\cite{piad-morffis_evolution_2015}. 
    \item \textbf{Step-Size Adaptation}: Supplementary to the default Cumulative Step size Adaptation (CSA), Two Point step size Adaption (TPA)~\cite{hansen_cma-es_2008} is implemented. TPA requires two additional objective function evaluations, used for evaluating both a shorter and a longer version of the population's center of mass. The version which shows the higher objective function value determines whether the step size should be increased or decreased. Five newly added mechanism for performing step size adaptation are implemented. They are described in detail in Section~\ref{subsec:ssa}. 
    \item \textbf{Mirrored Sampling}: For every newly sampled point, its mirror image is added the population, by reversing its sign~\cite{auger_mirrored_2011}. With \textit{Pairwise Selection}, only the best point of each mirrored pair is used in recombination.
    \item \textbf{Quasi-Gaussian Sampling}: Instead of performing the simple random sampling from the multivariate Gaussian, new solutions can alternatively be drawn from quasi-random sequences (a.k.a.~low-discrepancy sequences)~\cite{auger_algorithms_2005}. We implemented two options for this module, the Halton and Sobol sequences. 
    \item \textbf{Recombination Weights}: Three options are implemented; 1) default weights (see~\cite{cmatut}), 2) equal weights: $w_i=1/\mu$, and 3) $w_i = 1/2^{i} + 1/(\lambda2^\lambda)$ for $i=1,2,\ldots,\lambda$. 
    \item \textbf{Restart Strategy}: When the optimization process stagnates, the CMA-ES can be restarted using a restart strategy. Two strategies are implemented.  IPOP~\cite{auger_restart_2005} increases the size population after every restart by a constant factor. BIPOP~\cite{bipop} also changes the size of the population, but alternates between larger and smaller population sizes.
    \item \textbf{Boundary Correction}: If candidate solutions are sampled outside the search domain, they can be transformed back into the search domain by applying a boundary correction operation. In Section~\ref{sec:boundcorr}, we describe six options for performing boundary correction which have been implemented.
\end{enumerate}

In Table~\ref{tab:modules_options}, a detailed overview is given of all currently implemented modules and their options in the Modular CMA-ES framework.

\begin{table}[t]
\centering
\footnotesize
\begin{tabular}{llllllll}
\toprule
\# &  0 (default) & 1 & 2  & 3 & 4 & 5 & 6  \\ 
\midrule
1 &  off & on & - & - & - & - & -  \\
2 &  off & on & - & - & - & - & -  \\
3 &  off & on & - & - & - & - & - \\
4 &  off & on & - & - & - & - & -  \\
5 &  off & on & - & - & - & - & -  \\
6 &  CSA & TPA & \textbf{MSR} & \textbf{PSR} & \textbf{xNES} &\textbf{ m-xNES} & \textbf{p-xNES} \\
7 &  off & on & on w. PS & - & - & - & - \\
8 &  off & Sobol & Halton &  - & - & - & - \\
9 &  default & $\frac{1}{2^{i}} + \frac{1}{\lambda2^\lambda}$ & - & - & - & - \\
10 & off & IPOP & BIPOP &  - & - & - & -   \\ 
11 & off & \textbf{UR} & \textbf{MCS} & \textbf{COTN} & \textbf{SCS} & \textbf{TCS} & - \\
\bottomrule

\end{tabular}
\caption{The modules available for the Modular CMA-ES. The numeric index for each module corresponds to the index used in the text of Section~\ref{sec:modcma}. Newly added modules/options are given in bold.}\vspace{-10pt}

\label{tab:modules_options}
\end{table}

\subsection{Boundary Correction}\label{sec:boundcorr}
In the original framework, a boundary correction function taken from~\cite{ruili} was implemented, and always applied after each mutation. In some cases, however, this operator can degrade the performance of the algorithm quite drastically. We therefore decided to make the boundary correction operation optional, and to implement it as a module, for it to only be used when beneficial. A number of different boundary correction strategies were implemented, taken from~\cite{DBLP:journals/isci/CaraffiniKC19}:
\begin{enumerate}
    \item \textbf{None}: no correction is applied to infeasible coordinates of solutions.
    \item \textbf{Uniform Resample (UR)}: replaces all infeasible coordinates of a solution with new coordinates sampled uniformly at random within the search space.
    \item \textbf{Mirror Correction Strategy (MCS)}: mirrors all infeasible coordinates of a solution with respect to its closest boundary.
    \item \textbf{Complete One-tailed Normal Correction Strategy (COTN)}: All infeasible coordinates are replaces to new coordinates inside the search space according to a rescaled one-sided normal distribution centered on the boundary.%
    \item \textbf{Saturation Correction Strategy (SCS)}: All infeasible coordinates is set to the closest corresponding bound. 
    \item \textbf{Toroidal Correction Strategy (TCS)}: All infeasible coordinates get reflected off the opposite boundary.
\end{enumerate}

\subsection{Step-Size Adaptation} \label{subsec:ssa}
 In this work, we consider a number of alternative step size adaptation mechanisms for new options for the Modular CMA-ES. We take inspiration from~\cite{qualitative_ssa_krause_glas_2017}, which provides a qualitative evaluation of multiple step size adaptation mechanisms used in ES. In addition the CSA and TPA step size adaptation methods, which were already implemented, we implemented the following procedures:
\begin{enumerate}
\item{\textbf{Median success rule (MSR)}}: 
The MSR mechanism~\cite{msr} adapts the step-size $\sigma$ as follows: it firstly computes a success rate by checking the number of current individuals that are better than some user-defined quantile of the function values in the previous population, then accumulates such success rates in every iteration, and finally decides to increase the step-size if the cumulated values is bigger than $1/2$ and decrease it otherwise.

\item{\textbf{Population success rule (PSR)}}: determines the success rate of the current population using a rank-based approach. It firstly sorts all individuals in the current and previous population together, then retrieves the set of ranks of individuals belonging to the current iteration and the one for the previous iteration, and finally calculates the average rank difference between those two sets as the population success rate, which controls the step-size updates.

\item{\textbf{xNES step-size adaptation (xNES)}}: calculates the length of each standardized mutation vector 

and subtracts from it the expected length of the standard Gaussian vector. The resulting difference is then scalarized using the same weights used in the recombination, which is finally fed into an exponential function to generate a multiplicative coefficient to modify the step-size. 

\item{\textbf{mean-xNES step-size adaptation (m-XNES)}}: functions similarly to xNES, with the exception that it takes the standardized differential vector between current centers of mass and the one in the previous iteration and compares it to the expected length of the standard Gaussian vector. 

\item{\textbf{xNES with log normal prior Step size adaptation (p-xNES)}}: resembles the principle of self-adaptation for step-sizes, where $\lambda$ trial step-sizes are generated from a log-normal distribution which takes the current step-size as its mean and each trial step-size is used to sample a candidate point. To determine the new step-size, this method calculates the weighted sum of the log-transformed trial step-sizes, where those assigned to their corresponding candidate points in the recombination.
\end{enumerate}

\section{Incremental Assessment of Module Performance}

With the introduction of these new module settings, we have a clear use-case for the assessment of algorithmic ideas within the CMA-ES algorithm. Since these options are implemented into a framework with many existing modules, it will not suffice to look at them in isolation.  
Instead, we should carefully consider the potential interactions with the existing modules and investigate their impact on the empirical performance of ModCMA. Previous work~\cite{van_rijn_algorithm_2017} used data from a complete enumeration of all module settings to analyze the contribution of each individual module.
However, such an approach becomes intractable when we are confronted with a huge set of modules, or more importantly if we aim to obtain the contribution of some new modules implemented incrementally to an existing portfolio of modules, which we have investigated extensively.
Besides, this complete enumeration approach ignores entirely the configuration of continuous strategy parameters, e.g., $c_1$, $c_{\mu}$, and $c_c$, which have been shown to significantly impact the per-instance performance of the resulting configurations~\cite{BelkhirDSS17}.                 

To properly address the problem of determining the contribution of a single module setting to an existing portfolio of modules, we make use of hyperparameter optimization, which has previously been shown to achieve results comparable to the complete enumeration method, while being much more easily extendable to other hyperparameters~\cite{gecco_cash}. We propose the following roadmap to formalize this procedure:
\begin{enumerate}
    \item Select a modular implementation of the base algorithm to which the new module has been added, a hyperparamter optimizer and a performance metric.
    \item Collect a list of the existing modules and relevant hyperparameters (without the new module to assess). This will be the search space for the hyperparamter optimization.
    \item Run the selected hyperparameter optimizer on this search space, ideally for a wide set of relevant benchmark functions. This data will then serve as the baseline performance.
    \item Extend the original search space by including the new module to asses, and run the hyperparamter optimization on this extended search space (using the exact same setup as the baseline). 
    \item Compare the data from the baseline to the experiment with the extended search space. This should not only be done from a performance perspective, but also from the resulting configurations themselves. This allows for the analysis of potential interactions between modules. 
\end{enumerate}

\subsection{Performance Measures} \label{subsec:performance-measure}
In order to compare the different configurations of the ModCMA, we need to define the ways in which we measure their performance. 
Assuming a set of optimization algorithms $\mathcal{A}=\{A_1,A_2,\ldots\}$, a set of objective functions $\mathcal{F} = \{f_1,f_2,\ldots\}$, a function evaluation budget $B$, and $N$ repeated runs of each algorithm, we denote by $T(A, f, v, i)$, $i \in [1..N]$, the number of function evaluations consumed by algorithm $A$ to find in its $i$-th run on function $f$ a solution of solution quality at least $v$.  
Among various methods for quantifying the empirical performance of optimizers, the expected running time (ERT)~\cite{AugerH05} is commonly chosen, which estimates the expectation of the number of function evaluations (a.k.a.~the running time) of an optimizer to hit a predefined target value when unlimited evaluation budget is provided. However, the performance comparison based on ERT is be largely biased towards the target value prescribed by the user. This value can be difficult to determine a-priori for a configuration task on many optimizers and it also adds another design factor to our experimental setup. Instead, we propose to take a measure that relies on a \emph{set} of target values since it will be less sensitive to the choice of each individual value therein and could cover more perspectives of the running profile of optimizers. One of such measures is the the Area Under the ECDF Curve (AUC) of the running time, defined as follows:
\begin{align*}
    \AUC(A, f, \mathcal{V}) = \int_1^B \widehat{F}(t; A,f,\mathcal{V}) \mathrm{d} t,
\end{align*}
where $\widehat{F}(t; A,f, \mathcal{V}) = \frac{1}{N|\mathcal{V}|}\sum_{v\in \mathcal{V}}\sum_{i=1}^{N} \mathds{1}(T(A,f,v,i) \leq t)$ ($\mathds{1}$ is the characteristic function). In this work, we evaluate the algorithms for the target values $\mathcal{V} = \{10^{\frac{10-i}{5}} \colon i \in [1..51]\} \subset [10^{-8},10^2]$.

We note that most hyperparameter tuning methods are built with minimization in mind. As such, we use the Area Over the Curve (AOC) instead of AUC, since we know $\AOC(A, f, \mathcal{V})=B-\AUC(A, f, \mathcal{V})$.

\subsection{Technical Details}
The roadmap proposed above is designed to be generic, so that it can function with any modular algorithm, hyperparameter tuner, and performance metric. In order to collect the AOC measure from the runs of the ModCMA, we integrated it into the IOHprofiler~\cite{IOHprofiler}. This tool is used because it offers an easy way of accessing the BBOB-functions, while providing the needed logging functionality to easily calculate the AOC of each run. As our baseline, we will tune the existing modules from ModCMA, which are shown (plain text) in Table~\ref{tab:modules_options}, totalling 6 binary and 4 ternary modules. In addition, we tune the four continuous hyperparameters $c_1$, $c_\mu$, $c_\sigma$, $c_c$, and $c_\sigma$, which control the dynamics of the adaption of the covariance matrix ($c_1$, $c_{\mu}$, and $c_c$) and of the step-size ($c_\sigma$). 
We then run two experiments to assess both the new step-size adaptation methods and the boundary correction module, as introduced in Section~\ref{subsec:ssa} and Section~\ref{sec:boundcorr} respectively. 
All of the code used in these experiments, and the resulting data, is available in~\cite{code}. 

\textbf{Hyperparameter tuning using irace:} 
In this paper, we use the irace~\cite{irace, irace_2011} library as our hyperparameter optimizer. Irace\footnote{Implemented in R, freely available at~\cite{irace_code}.} is based on the principle of iterated racing, in which each race\footnote{The initial iteration of irace consists of random configurations and the default CMA-ES setting.} repeatedly executes configurations on different problem instances until there is statistically significant reason to discard enough of them to move to the next race (thus inherently allocating more runs to more promising configurations). Using this procedure, one or more configurations will emerge as the final elites at the end of the optimization. The number of times irace has evaluated the elite configurations can differ significantly between two runs. To obtain a fair comparison, we therefore 
perform an independent set of 25 validation runs, with the same random seeds for all configurations. We use the results of these runs to assess the final performance. 

\textbf{BBOB problem suite:} 
We configure irace to use the first instance (\texttt{iid = 1}) of each of the 24 BBOB functions~\cite{bbobfunctions,hansen2020coco}, in 5D. While the argument can be made that tuning should be done over multiple instances, we favoured running more repetitions of irace over using more instances. Each irace-run is given a budget of 1\,000 algorithm evaluations, which themselves have a budget of $10\,000\cdot D$ function evaluations. 

\section{Results}
\begin{figure}
    \centering
    \includegraphics[width=0.5\textwidth, trim={0, 0, 0, 0},clip]{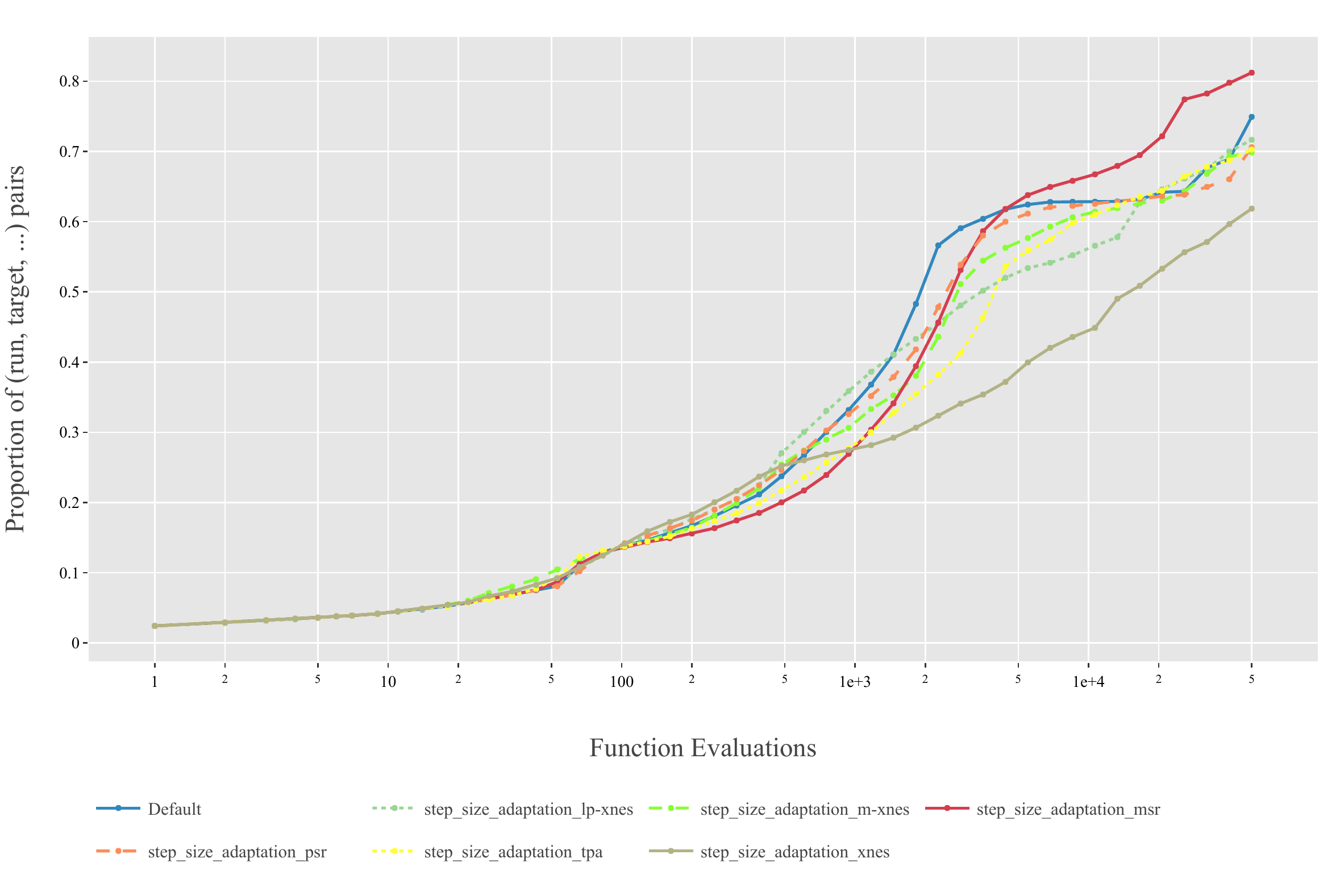}
    \caption{ECDF-curve of all single-module stepsize options. Figure generated using IOHanalyzer~\cite{IOHanalyzer}.}\vspace{-15pt} 
    \label{fig:ecdf_single_mod}
\end{figure}

Before considering our proposed method, we run a basic benchmarking experiment on each of the individual module options. This is similar to the common approach of benchmarking the new module against a set of other algorithm variants. We show the resulting best single-module configurations (a.k.a.~the virtual best solver, VBS for short) relative to the default CMA-ES in Table~\ref{tab:VBS_single_module}. In this table, we see that among the new modules, only two have been selected: MSR for F23 and m-XNES for F5. We can further look at the over-all contributions of the newly introduced step-size settings by plotting the ECDF-curves over all functions, as done in Figure~\ref{fig:ecdf_single_mod}. In this figure, we can clearly see that most methods are quite competitive, with the only exception being xNES, which has a significantly worse performance than the others. Overall, the MSR method seems to be quite effective, but there is no strict domination over the other settings. 

\begin{table}[!ht]
    \scriptsize
    \centering
\begin{tabular}{rlrrr}
\toprule
 Fid &                          VBS & AOC of VBS & AOC of Default & Improvement \\
\midrule
	1	&	elitist\_True	&	247	&	326	&	24\%	\\
	2	&	active\_True	&	1\,272	&	1\,659	&	23\%	\\
	3	&	local\_restart\_BIPOP	&	38\,374	&	44\,518	&	14\%	\\
	4	&	local\_restart\_IPOP	&	41\,746	&	44\,613	&	6\%	\\
	5	&	step\_size\_adaptation\_m-xnes	&	43	&	63	&	31\%	\\
	6	&	elitist\_True	&	655	&	904	&	28\%	\\
	7	&	step\_size\_adaptation\_tpa	&	1\,312	&	39\,199	&	97\%	\\
	8	&	base\_sampler\_halton	&	1\,186	&	4\,544	&	74\%	\\
	9	&	base\_sampler\_sobol	&	959	&	2\,470	&	61\%	\\
	10	&	active\_True	&	1\,309	&	1\,729	&	24\%	\\
	11	&	active\_True	&	1\,162	&	1\,749	&	34\%	\\
	12	&	base\_sampler\_sobol	&	2\,186	&	2\,980	&	27\%	\\
	13	&	active\_True	&	1\,627	&	2\,191	&	26\%	\\
	14	&	active\_True	&	601	&	831	&	28\%	\\
	15	&	local\_restart\_BIPOP	&	30\,380	&	43\,313	&	30\%	\\
	16	&	local\_restart\_BIPOP	&	8\,172	&	34\,132	&	76\%	\\
	17	&	threshold\_convergence\_True	&	12\,464	&	26884	&	54\%	\\
	18	&	threshold\_convergence\_True	&	15\,764	&	33724	&	53\%	\\
	19	&	mirrored\_mirrored	&	33567	&	36\,688	&	9\%	\\
	20	&	threshold\_convergence\_True	&	36\,482	&	40691	&	10\%	\\
	21	&	local\_restart\_IPOP	&	38\,028	&	40\,371	&	6\%	\\
	22	&	mirrored\_mirrored	&	566	&	8\,632	&	93\%	\\
	23	&	step\_size\_adaptation\_msr	&	11\,060	&	34\,433	&	68\%	\\
	24	&	local\_restart\_IPOP	&	42\,099	&	44\,351	&	5\%	\\

\bottomrule
\end{tabular}
    \caption{Table showing the AOC of the best single-module configuration for each function (VBS), compared to that of the default CMA-ES. 
    Note that these values does not include benefits from tuning the continuous hyperparameters, which are set to the default values for all configurations in this table.\vspace{-10pt}  
    }
    \label{tab:VBS_single_module}
\end{table}

\subsection{Baseline}
\begin{figure}[t]
    \centering
    \includegraphics[width=0.5\textwidth, trim={100, 30, 10, 50},clip]{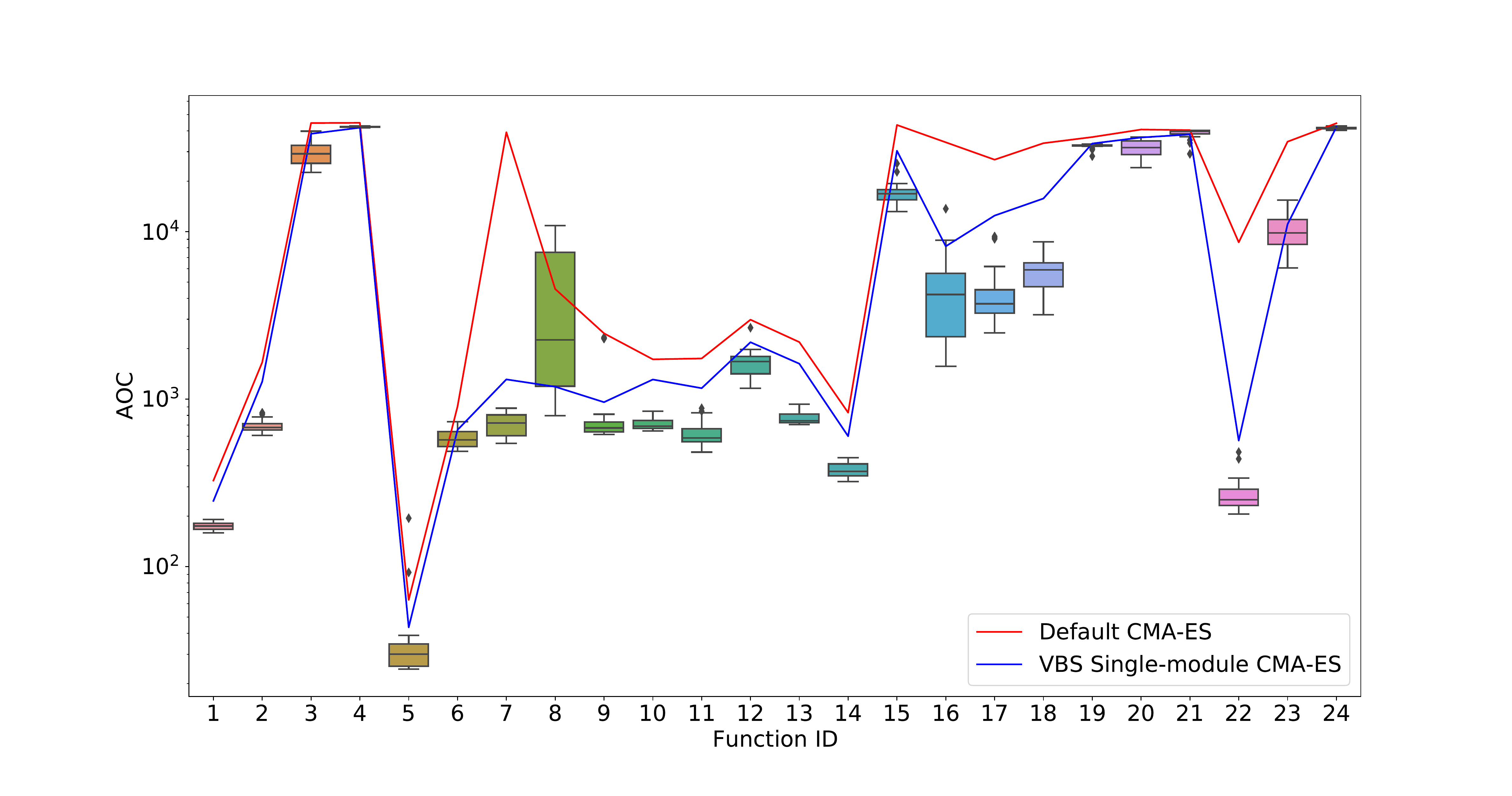}
    \caption{Distribution of the area over the ECDF curve for the final elite configuration of the baseline irace runs. All AOC's are averages of 25 verification runs. The VBS single-module configurations can be seen in Table~\ref{tab:VBS_single_module}. }
    \label{fig:boxplot_baseline}
\end{figure}

To illustrate the usefulness of hyperparameter tuning in the modular CMA-ES, we conduct a baseline experiment where we tune all modules (excluding the newly introduced ones) and the selected hyperparameters. We perform this tuning on each benchmark function separately
, and compare this baseline to the default CMA-ES as well as the virtual best solver on each function from Table~\ref{tab:VBS_single_module}.
Since we run 4 runs of irace for each function, this results in 4 sets of elites (each set has up to 5 configurations), for which we then perform the verification runs. We plot the distribution of the AOC for each of these configurations in Figure~\ref{fig:boxplot_baseline}. From this figure, it is clear that the tuning of all parameters at once is much better than simply selecting a single-module variant, as is to be expected. This plot also highlights the significant differences in performance of the final found configurations. There are two main reasons for this fact: the inherent stochasticity of the CMA-ES itself, and the large impact of the initially generated configurations of irace. We discuss these challenges in detail in Section~\ref{sec:challenges}.

\begin{figure}[!t]
    \centering
    \includegraphics[width=0.5\textwidth, trim={100, 0, 10, 50},clip]{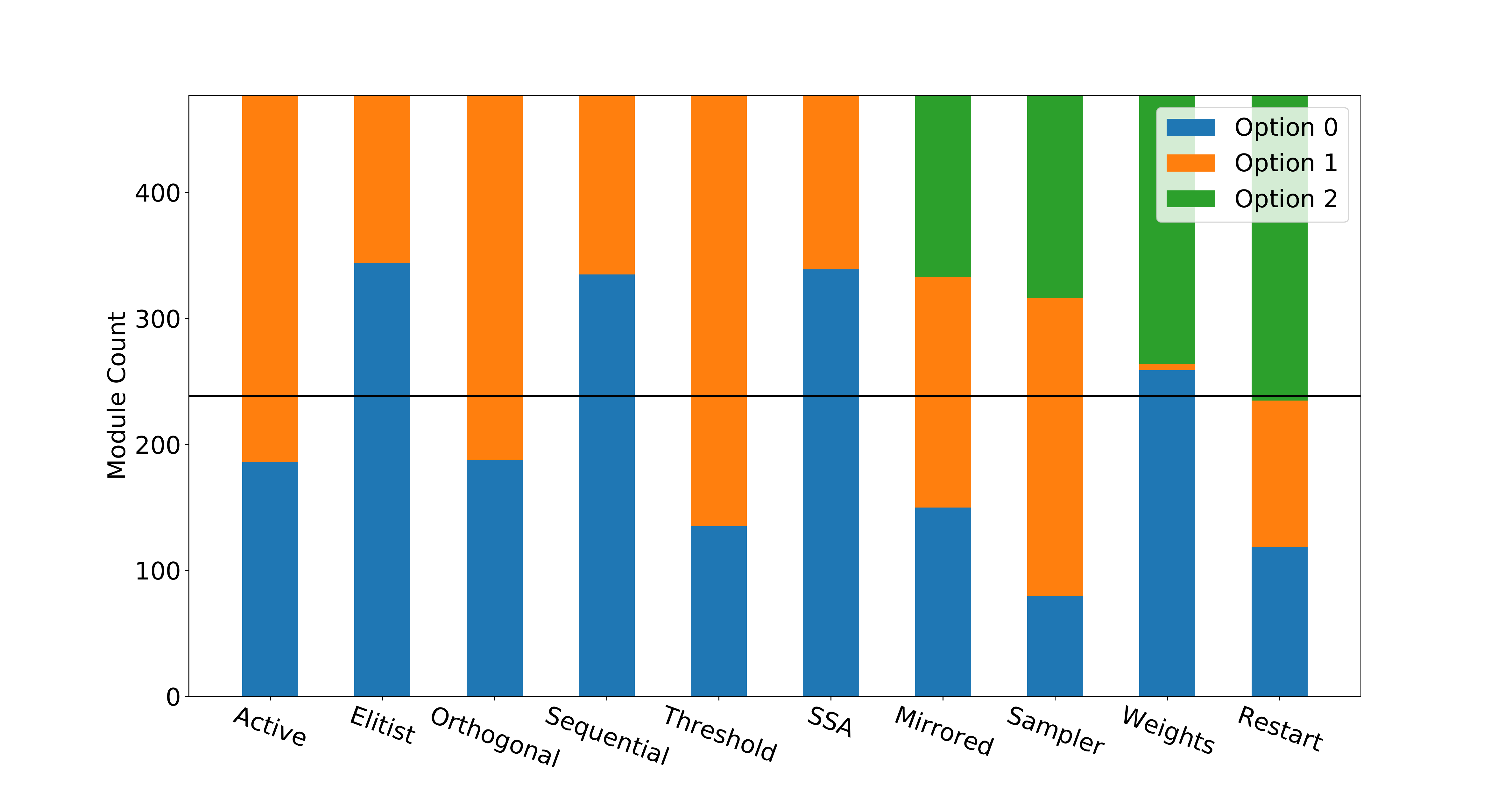}
    \caption{Module counts of all elites found in the baseline-experiment, over all 24 BBOB-functions. The option numbers correspond to those in Table~\ref{tab:modules_options}}
    \label{fig:stacked_bars_base}
\end{figure}

From this baseline data, we can also study the resulting configurations themselves. This can be done by aggregating the modules which have been selected in the final elite configurations in the separate irace runs, as is visualized in Figure~\ref{fig:stacked_bars_base}. In this figure, we can see that there is a large variability in the selected module options, which seems to indicate that they are all usable for at least some functions. One notable exception is the weights option ``equal'', which is chosen in less than $1\%$ of configurations. 

\begin{figure}
    \centering
    \includegraphics[width=0.5\textwidth, trim={100, 30, 10, 50},clip]{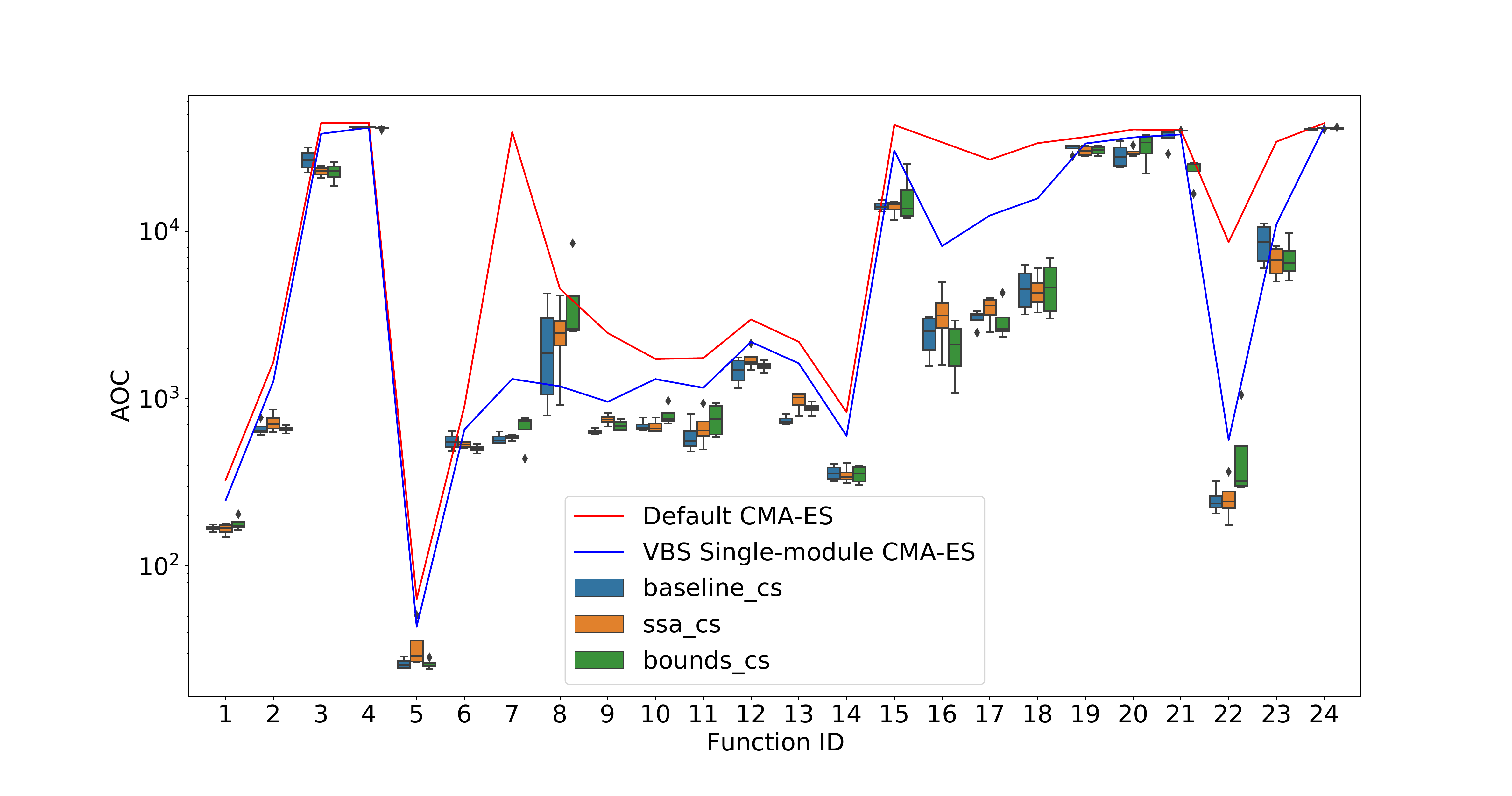}
    \caption{Distribution of the single best elites from the baseline and the tuning with the additional modules. AOC values are the result of averaging over 25 verification runs.}
    \label{fig:all_perf_plot}
\end{figure}

\subsection{Performance analysis}\label{subsec:performance-analysis} 
For generating our experimental data, we conduct two hyperparameter optimization experiments with irace, one where allow the new SSA methods to be selected, and another including the new boundary correction methods. Note that in the boundary correction experiment, the new SSA methods cannot be selected and vice versa. We use the same experimental setup for running these experiments as with the baseline experiment.

Based on these experiments, there are two main approaches to analyze the contributions of the newly introduced modules: the performance-perspective and the perspective of the selected modules. We start by looking at the performance: for each experiment, we look at the impact 
on the final performance of the elite configurations found by the irace runs. First, we visualize the distributions of the AOC of the single best configuration found in each run of irace (based on the verification runs) in Figure~\ref{fig:all_perf_plot}. 

In this plot, we can see that the effect of introducing the new modules is quite mixed. For some functions, the performance even worsens significantly (e.g., on F8) after introducing new modules, while for others we see the desired improvement (e.g. on F23)
\begin{figure}[t]
    \centering
    \includegraphics[width=0.5\textwidth, trim={0, 0, 0, 0},clip]{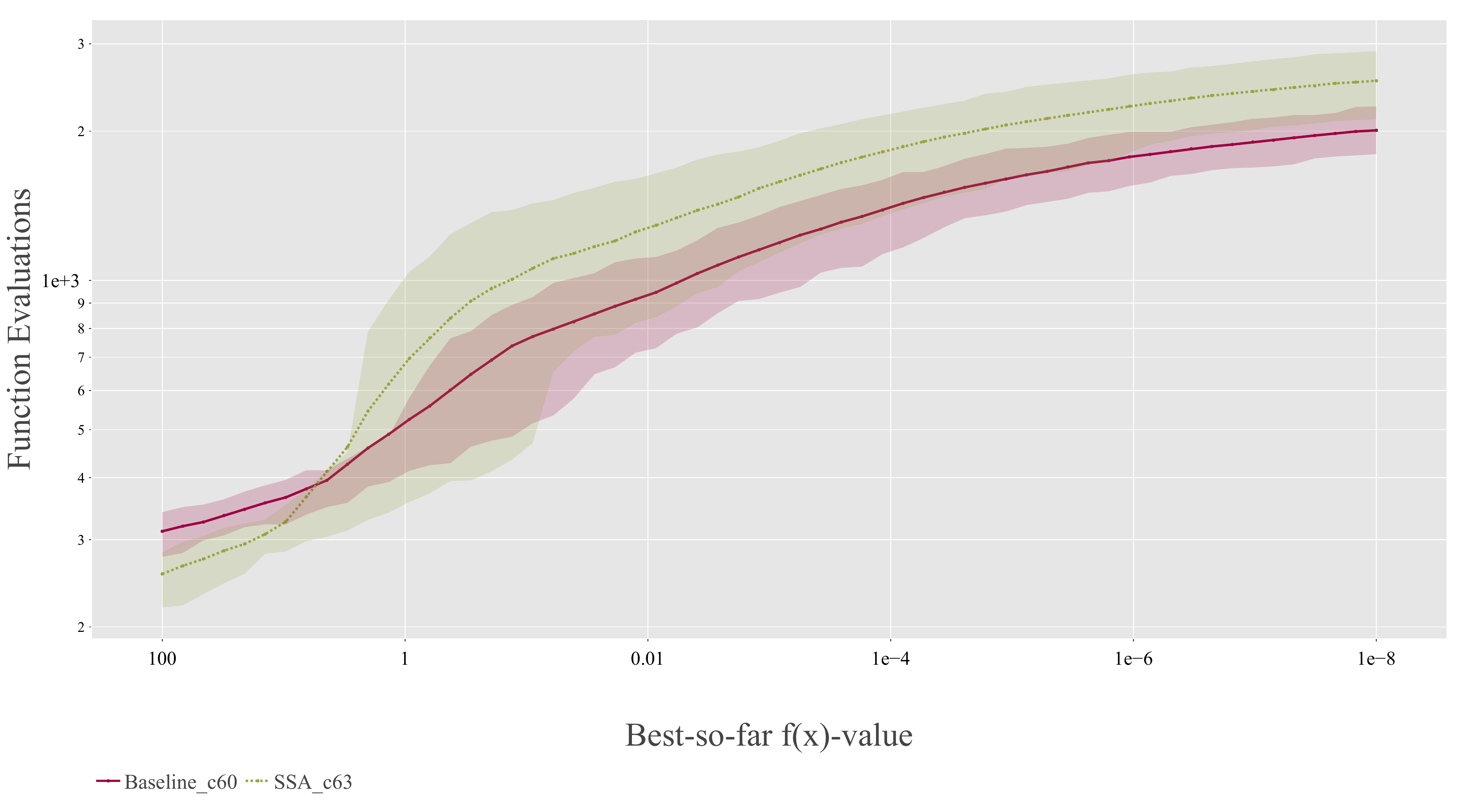}
    \caption{Comparison of the Expected Running Time of the best configurations found on F12 by both the baseline and the SSA experiments. Shaded areas indicate the outer quantiles (20-80).}\vspace{-10pt}
    \label{fig:F12_ERT}
\end{figure}

\begin{figure}[t]
    \centering
    \includegraphics[width=0.5\textwidth, trim={70, 40, 10, 60},clip]{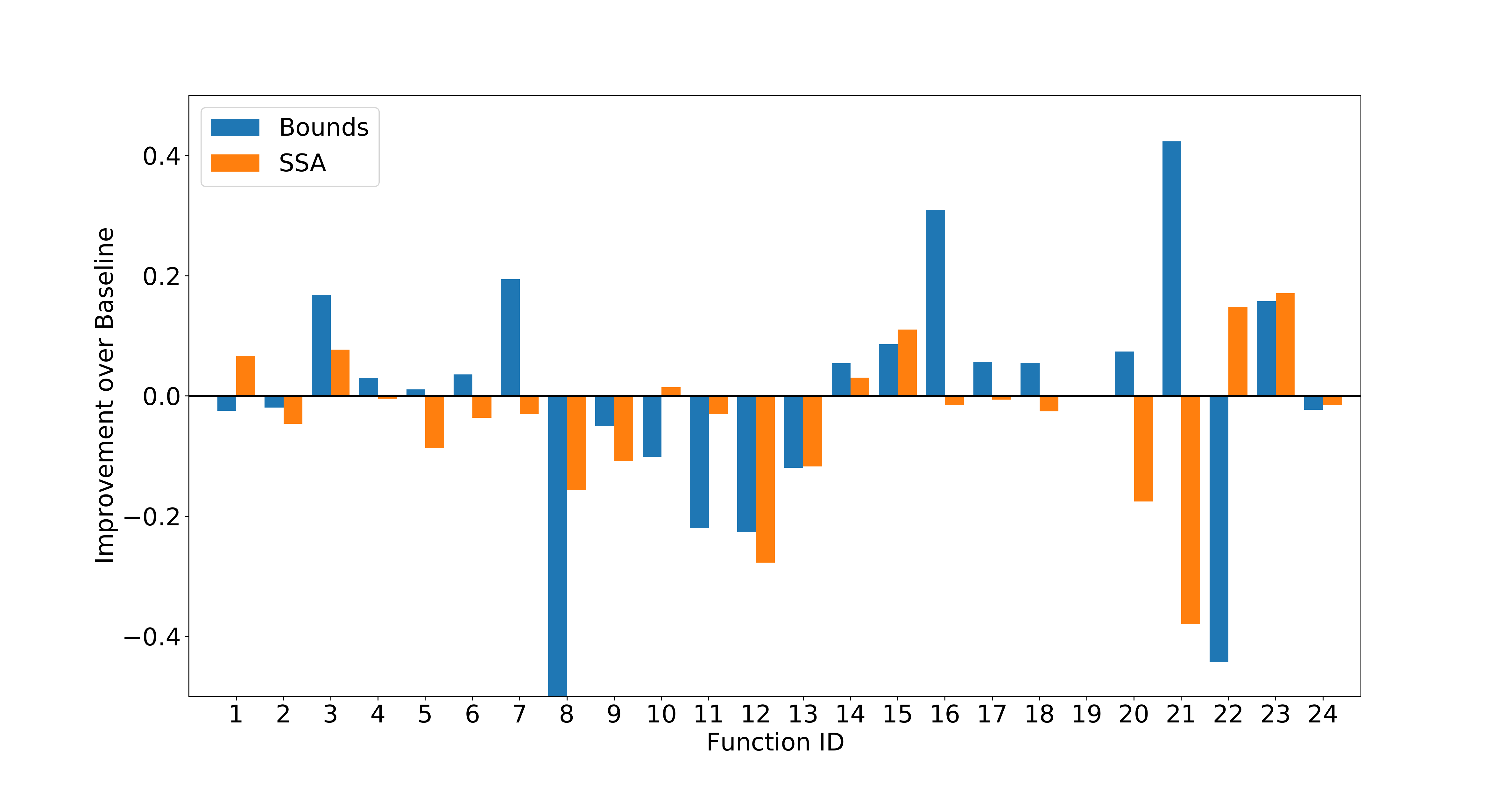}
    \caption{The relative improvements per function of the best configuration found by irace relative to the baseline experiment's best configuration. }
    \label{fig:rel_impr_barplot}
\end{figure}

In order to better show these differences, we plot in Figure~\ref{fig:rel_impr_barplot} the AOC from the single best configurations found in both the SSA and bound-experiments relative to the best configuration from the baseline. 
We see that the general trend of performance is somewhat negative, with some outliers in both directions. This seems to indicate that these new modules are not always beneficial to the final performance. For example, we can consider F12, where the configuration found by the baseline has an average AOC of $1\,159$, while the best configuration found when including the new SSA-methods in the search space reaches an average AOC of $1\,480$. We show the expected running time of thes two configurations in Figure~\ref{fig:F12_ERT}, where we can clearly see this difference. However, it is also clear that the variance in between runs 
is significant, which can partly explain poor performance. Indeed, if we look at the average AOC during the irace runs, the difference between these two configurations is only $7\%$. This leads to an important observation about the assessment of the new algorithmic modules: when judging results purely from the average performance measures, it is necessary to also consider the overall variability of the experiment, as well as the inherent stochasticity of the base algorithm. 

We perform the same procedure to the boundary correction methods. The impact of this module is expected to be smaller, since for most of the ``easier'' functions, the boundary condition is rarely violated. For some of the more challenging functions however, the penalty value given by BBOB function itself might not be sufficient to ``guide'' the algorithm back in bounds, but an explicit boundary correction could be beneficial in these cases. We can see that this seems to indeed be the case in Figure~\ref{fig:rel_impr_barplot}, where on the more complex functions, e.g., F21, the performance is improved when the boundary correction module is tuned. We also see that for these functions, the ``None'' option is rarely selected, which confirms that the algorithm jumping out of bounds without being corrected has negative impact on the performance.

In Figure~\ref{fig:rel_impr_barplot}, we also see that, the inclusion of the new step-size adaptation methods does manage to improve the overall performance for some functions. As an example, on F23 we saw an improvement of $17.1\%$ over the best baseline configuration. If we consider all four elite configurations and compare the average performance differences, the average improvement is even higher, at $22.3\%$. The stability of this improvement is promising, but in order to fully grasp how the inclusion of the new step-size adaptation mechanisms leads to this improvement, we need to analyze the selected modules across these different experiments.

\subsection{Module Analysis}
\begin{figure}[t]
    \centering
    \includegraphics[width=0.5\textwidth, trim={100, 20, 10, 70},clip]{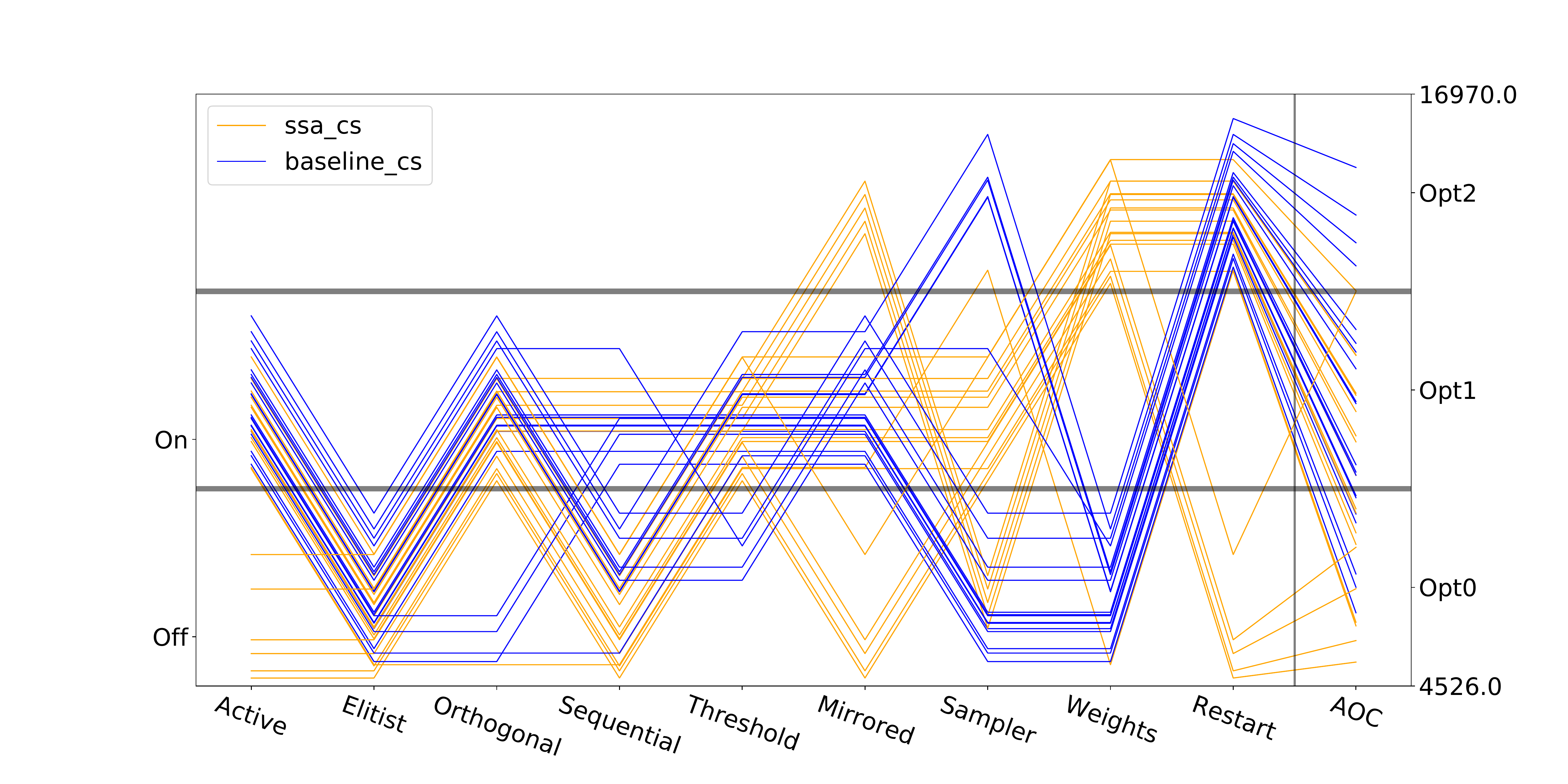}
    \caption{Combined module activation plot for the elites found in the baseline and SSA experiments, for function 23. The lower the line, the better its performance, scaled within each band according to the AOC. The option numbers correspond to those in Table~\ref{tab:modules_options}.}
    \label{fig:mod_act_F23}
\end{figure}

We have seen that the performance of the elite configuration found on F23 improves when we include the new step-size adaptation modules in the search space. In order to identify what this performance can tell us about the new modules themselves, we should study the configurations in more detail. The obvious way to see the difference is by looking at how often the new module options have been selected in the final elite configurations. Over 20 elites, the PSR update was selected 14 times, MSR once, and CSA five times. This shows that these new modules are indeed used in the successful configurations. To see how the inclusion of these module options changes the interactions with the other modules, we look at the combined module activation plot, which is shown in Figure~\ref{fig:mod_act_F23}.
From this figure, we can see that there are some interesting differences between the two sets of configurations: the options for the restart and mirrored module are not as uniform when using the new step-size adaptation methods, and the weights option is changed completely. 
These observations show that there is a clear interplay between these modules.

\begin{figure}
    \centering
    \includegraphics[width=0.5\textwidth, trim={0, 5, 10, 20},clip]{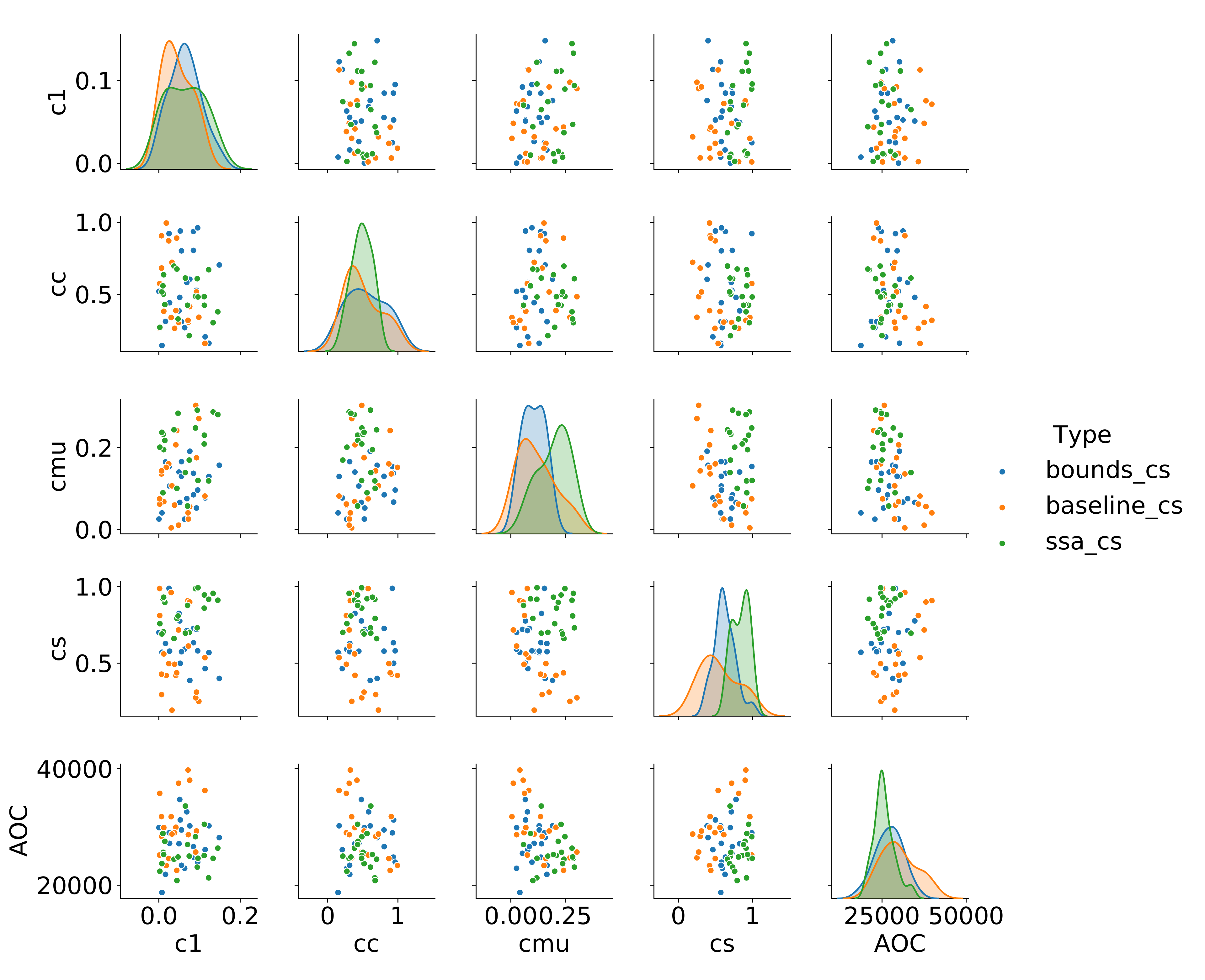}
    \caption{Distribution of the continuous hyperparameters from the elite configurations found in all three experiments.}
    \label{fig:pairplot_cont_F3}
\end{figure}

Next to the module activations, we can also look into the distributions of configured continuous hyperparameters. To illustrate this, we study F3, and plot the pairwise relations between the four continuous hyperparamters and the final AOC value in Figure~\ref{fig:pairplot_cont_F3}. From the marginal distribution (shown on the diagonal), we can see that the optimized setting of $c_\sigma$ differs the most across the SSA, boundary correction, and the baseline experiments.
This is a direct result of the introduction of the new step-size adaptation methods, each of which prefers slightly different settings for this parameter. This indicates that even though the final performance of the elite configurations is similar between the baseline and the SSA-experiment, the inclusion of new step-size adaptation methods significantly alters the found elite configurations. 

\begin{figure}
    \centering
    \includegraphics[width=0.5\textwidth, trim={0, 0, 10, 50},clip]{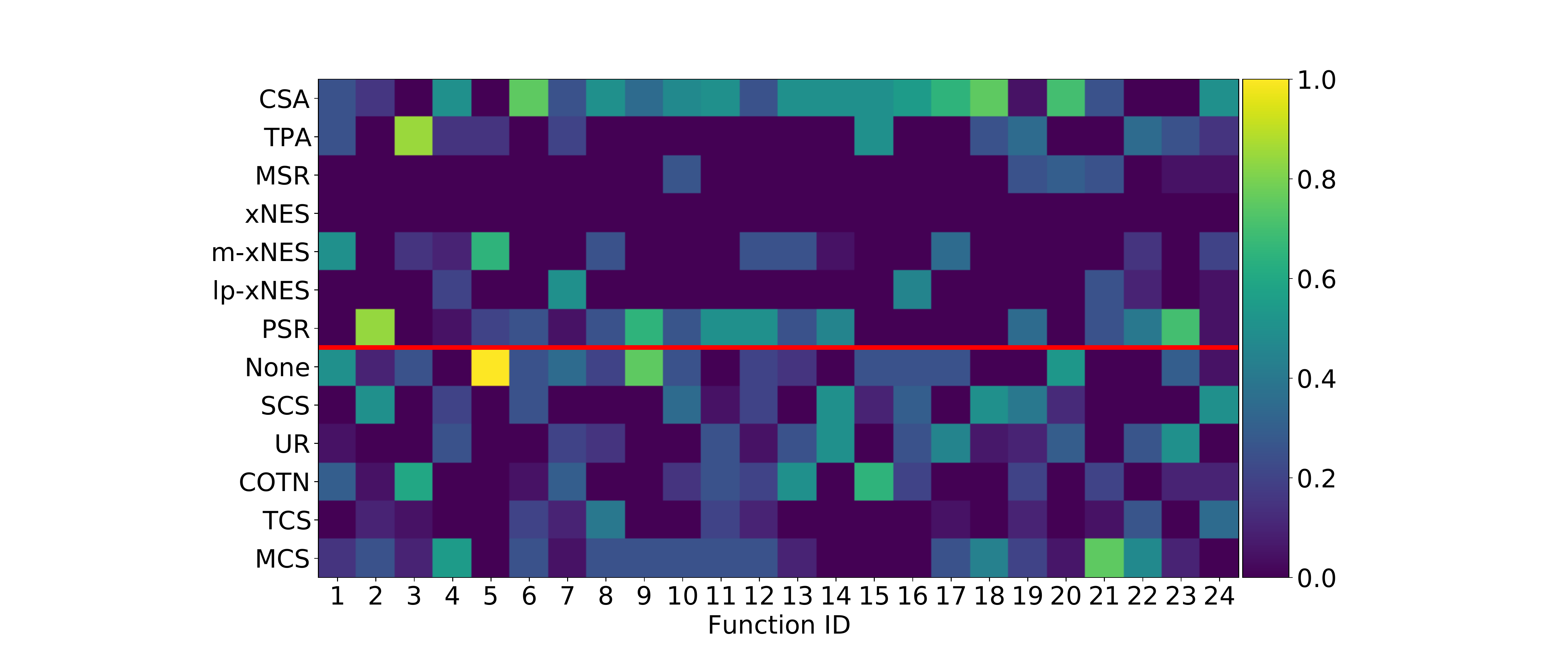}
    \caption{Heatmap showing the fraction of the elite configuration in which each of the options for either SSA (top) or boundary correction (bottom) are active.\vspace{-10pt}}
    \label{fig:mods_new}
\end{figure}

\begin{figure}
    \centering
    \includegraphics[width=0.5\textwidth, trim={0, 50, 10, 200},clip]{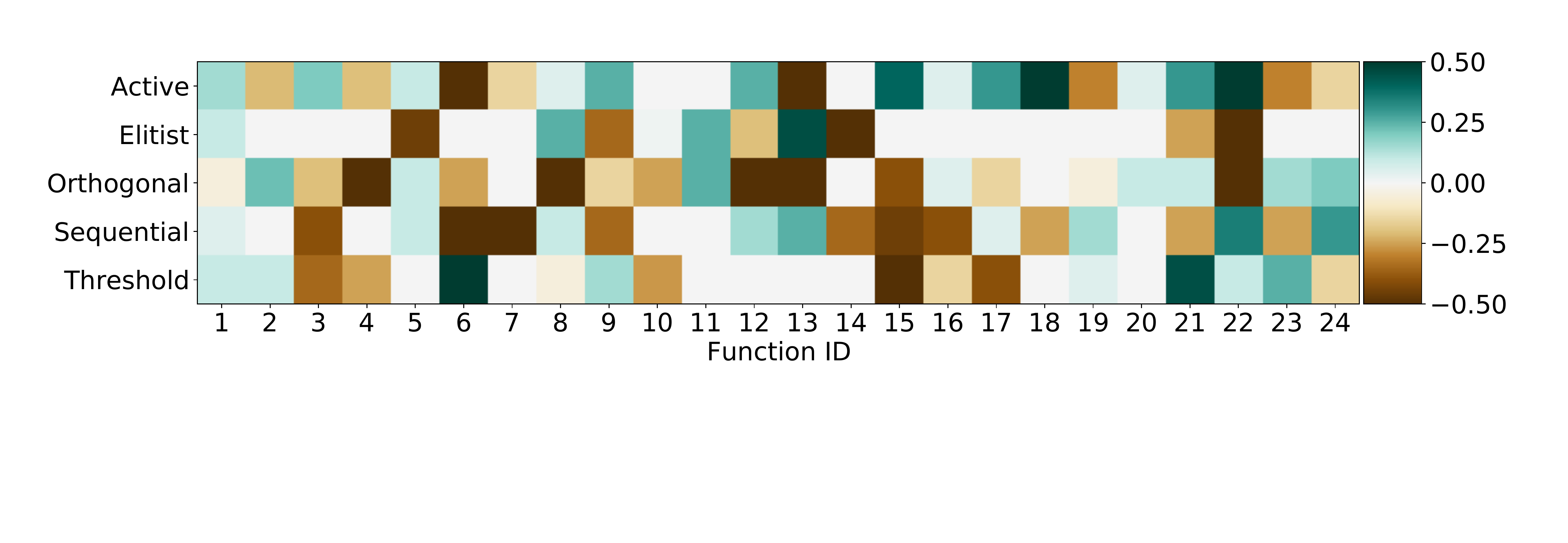}\vspace{-30pt}
    \caption{Heatmap showing difference in the fraction of the elite configuration in which each the of the binary modules are active, between the baseline and the SSA experiment. Positive values indicate a module is turned on more often in the SSA experiments.}
    \label{fig:heatmap_baseline}
\end{figure}

\begin{figure}
    \centering
    \includegraphics[width=0.5\textwidth, trim={0, 50, 10, 200},clip]{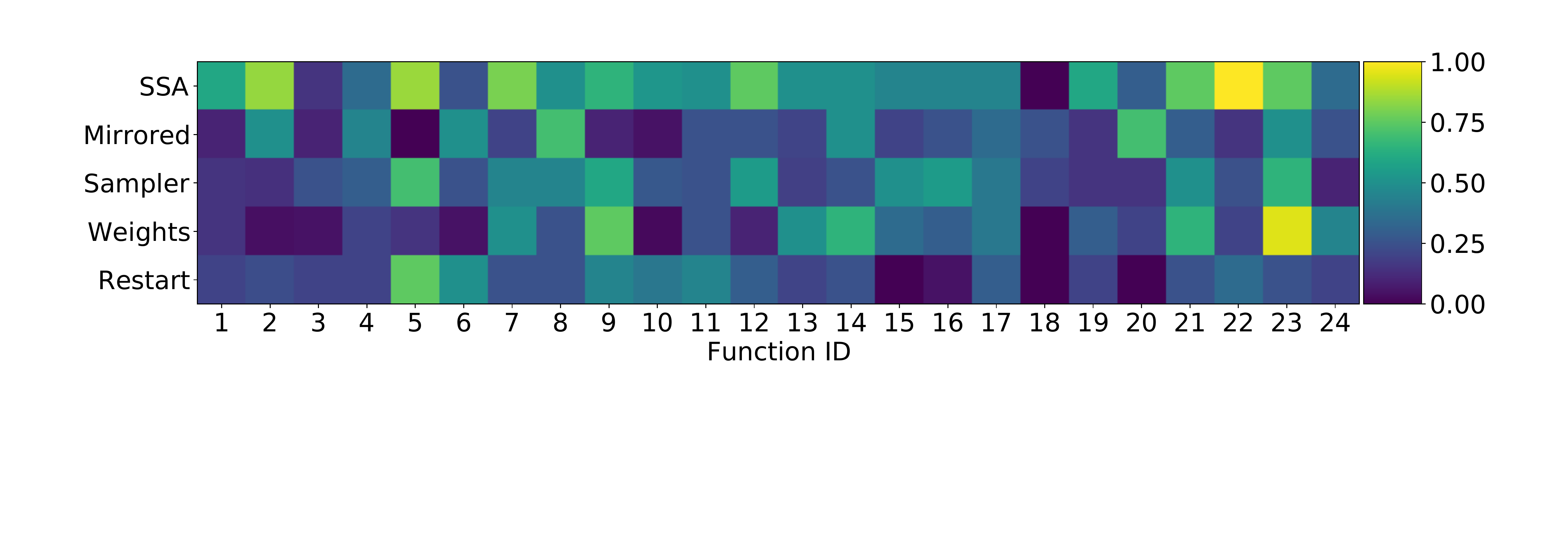}\vspace{-30pt}
    \caption{Heatmap showing difference in the distribution of the ternary modules selected in the final elites, between the baseline and the SSA experiment. 0 indicates that the distribution is identical, while 1 indicates that there is no overlap at all.}
    \label{fig:heatmap_baseline_ternary}
\end{figure}

We can extend this module analysis to all functions by aggregating the most important differences found between the baseline and SSA-experiments. First, we can plot how often each new module option is selected in the elites for each function, as is done in Figure~\ref{fig:mods_new}. We can use the same principle to study the interaction with the other modules. For the binary modules, we can directly capture the module difference by looking at which modules occur more or less often in the final set of elites, as is visualized in Figure~\ref{fig:heatmap_baseline}. While this does not directly generalize to modules with more settings, we can create a similar plot for the other modules by considering the overlap in selected module distributions. This is visualized in Figure~\ref{fig:heatmap_baseline_ternary}. From these figures, it becomes clear that the elites on some functions are barely affected by the inclusion of the new modules, while others require completely different module settings to properly exploit the changed search space. 

\begin{figure}
    \centering
    \includegraphics[width=0.5\textwidth, trim={100, 30, 10, 50},clip]{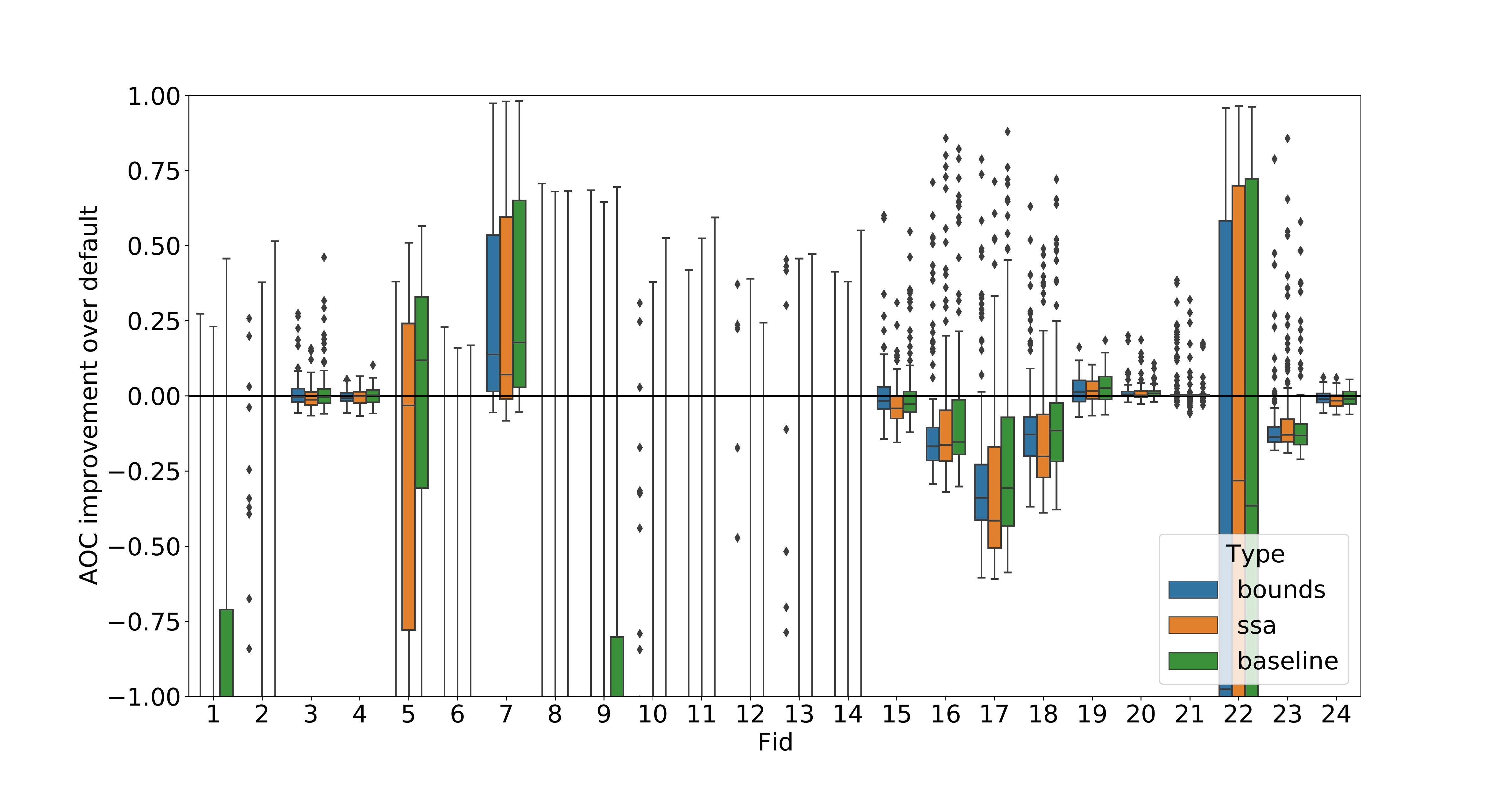}
    \caption{Distribution of the relative AOC values found in the initial race of irace (relative to the default CMA-ES configuration; positive values equate to lower AOC.)}\vspace{-20pt}
    \label{fig:initial_conf_perf_relative}
\end{figure}

We should note that only considering the final elite configurations does not tell the full story of a modules contribution. As briefly noted previously, introducing a new module increases the size and complexity of the search space, which has a large impact on the hyperparameter tuning. If a module is very dependent on the settings of other hyperparameters, this can lead to deterioration of the final results, since the initially sampled configurations are likely to have worse performance than those in the baseline. This is visualized in Figure~\ref{fig:initial_conf_perf_relative}, where this is clearly seen on function F5. This is a linear slope function, but the BBOB-specification does not include a sufficient penalty for leaving the search space. As a result, an algorithm which quickly leaves the search space will reach the required objective value very quickly. Thus, when adding boundary correction methods, $5/6$ random configurations are not able to abuse this loophole, leading to a worse initial performance. While for F5, the function is simple
enough that the good configurations can still be found (and the inclusion of the default CMA-ES settings in the initial population means that there is always at least one good configuration present),
the same issue exists to a lesser extent in other functions. Figure~\ref{fig:initial_conf_perf_relative} also shows that the ``tunability'' of modules on different functions varies widely. For instance, on functions F16 - F18, the spread of AOC values are significantly larger than those on functions F19 - F21, suggesting that it is relatively more difficult to tune the modules in the latter since the tuner will very likely take a considerably larger budget to identify optimal configurations. Also, while on some functions it is trivial to get improvement (e.g., F7) over the default CMA-ES, it is a lot more challenging on others, for example on functions F16 - F18.

\section{Challenges}\label{sec:challenges}
We discuss three key challenges for the module assessment procedure based on on hyperparameter optimization that we have identified in this work. 

\textbf{Influence and stochasticity of the hyperparameter tuning:} While we showed that assessing the impact of an algorithmic component by using a hyperparameter tuning approach provides useful insights, there are several factors which can complicate this approach. Since hyerparameter tuning is a very challenging problem, with many different approaches to solving it, the kind of tuner used will have a large impact on the resulting assessment. In this paper, we used irace, which tends to focus on converging to a single configuration, instead of covering a large set of different solutions. This necessitates running multiple repetitions of the irace procedure itself, as the initialization might otherwise have too much impact on the final configurations. This can quickly become computationally expensive.

\textbf{Algorithm-inherent stochasticity:} 
As we discussed in the results, we need to take care when drawing conclusions from the performance of the different CMA-ES configurations. Since CMA-ES is inherently stochastic, the amount of variance of the configurations on a certain function has a large impact on the search procedure of irace. Since we end up selecting elites based on the average performance, we are inherently underestimating the AOC of the final configuration. Even though irace largely mitigates this by using statistical testing in the races to decide when to discard configurations, there will always be some degree of underestimation of the performance (the median performance in the verification runs is $3.4\%$ worse than predicted from the irace runs). 
\textbf{Limits of the per-instance analysis:} 
In the current setup, the performance measures are only done on an per-instance basis. While this is often preferred over tuning for large sets of functions/instances, it does have some drawbacks. Specifically, if a module is designed to have a good performance over a wide set of functions, but other settings exist for each individual function which outperform it, this new module would not be seen as beneficial. Because of this, we argue that module assessment by hyperparameter tuning should not replace the traditional assessments, but rather complement it for more in-depth, per-instance analysis. We can identify this for the step-size adaptation module by looking at the ECDF-curve of the single-module variants, as previously shown in Figure~\ref{fig:ecdf_single_mod}. To assess the impact of a new algorithmic component in a robust manner, tuning \emph{across} a whole benchmark set of possibly diverse problems can be performed, and compared to a tuned variant of the same modular framework without this module.

\section{Discussion and Future Work}
We introduced a roadmap for assessing the performance of individual algorithmic ideas, which takes into account the interplay with other existing settings by comparing the results of hyperparameter tuning. Since this approach requires a modular design to function as intended, we use the Modular CMA-ES framework, which we have extended with 
 new modules. Our analysis showed that the newly added step size adaptation mechanisms are not always useful, but do provide clear benefits in several functions. The results also showed that step-size adaptation is most useful when combined with a different weights option.

The current version of the Modular CMA-ES framework is a good step in the direction of complete modularization of the CMA-ES algorithm, but some further enhancements can still be made. This would allow for even more precise control over each of the individual components, leading to an ideal testbed for new algorithmic ideas, which can then be evaluated using the approach outlined in this paper. However, since this can be computationally intensive, we should aim to share and reuse data as much as possible, by developing and maintaining a well-organized repository for this type of benchmark data. This does not only reduce the amount of computation needed to test new modules, but it also gives rise to the possibility of testing methods to re-use data from other experiments, since the search spaces have large overlap. Ideally, this would allow for the usage of methods from transfer learning to significantly shorten the time needed to assess a modules performance, even within a large modular search space.  

Additionally, we note that while the proposed module assessment is inherently dependent on the used hyperparameter tuning method, the overall procedure remains the same no matter which tuner is used. As a result, the analysis of the results should take into account the particularities of the tuner, such as the way configurations are generated. Further research should still be done into different hyperparameter optimization methods (e.g., SMAC~\cite{SMAC}, MIP-EGO~\cite{MIP-EGO}, SPOT~\cite{SPOT}, GGA~\cite{GGA}, hyperband~\cite{li2016hyperband}, etc.) to determine exactly how they differ in this modular algorithm context. Additionally, an analysis pipeline for this type of benchmarking could be designed within existing tools like the IOHanalyzer~\cite{IOHanalyzer}, which would greatly reduce the amount of effort needed to assess new algorithmic ideas.

\subsection*{Acknowledgements}
This work was supported by the Paris Ile-de-France Region. 

\bibliographystyle{ACM-Reference-Format}
\bibliography{references}

\end{document}